\documentclass{article}
\usepackage{ijcai22}

\usepackage{times}
\usepackage{soul}
\usepackage{url}
\usepackage[hidelinks]{hyperref}
\usepackage[utf8]{inputenc}
\usepackage[small]{caption}
\usepackage{graphicx}
\usepackage{amsmath}
\usepackage{amsthm}
\usepackage{booktabs}
\usepackage{algorithm}
\usepackage{algorithmic}
\title{Building a visual semantics aware object hierarchy \\ {\normalsize PhD Proposal}}

\author{
    Xiaolei Diao
    \affiliations
    DISI, University of Trento, Italy
    \emails
    xiaolei.diao@unitn.it
}

\begin{document}

\maketitle

\begin{abstract}
The semantic gap is defined as the difference between the linguistic representations of the same concept, which usually leads to misunderstanding between individuals with different knowledge backgrounds. Since linguistically annotated images are extensively used for training machine learning models, semantic gap problem (SGP) also results in inevitable bias on image annotations and further leads to poor performance on current computer vision tasks.
To address this problem, we propose a novel unsupervised method to build visual semantics aware object hierarchy, aiming to get a classification model by learning from pure-visual information and to dissipate the bias of linguistic representations caused by SGP. 
Our intuition in this paper comes from real-world knowledge representation where concepts are hierarchically organized, and each concept can be described by a set of features rather than a linguistic annotation, namely visual semantic. 
The evaluation consists of two parts, firstly we apply the constructed hierarchy on the object recognition task and then we compare our visual hierarchy and existing lexical hierarchies to show the validity of our method. The preliminary results reveal the efficiency and potential of our proposed method.
\end{abstract}

\section{Introduction}
\label{sec:Introduction}
Technology in the Machine Learning (ML) area is being widely applied in more and more real-world tasks, especially in the computer vision area, where the boom in neural networks has brought disruptive innovations that allow computers to perceive the world and complete a series of tasks. 
As a data-driven science, how the performance of ML models is implicated by the quality of datasets has received increasing concern \cite{1}.

\begin{figure}[tbp]
	\centering
	\includegraphics[scale=0.28]{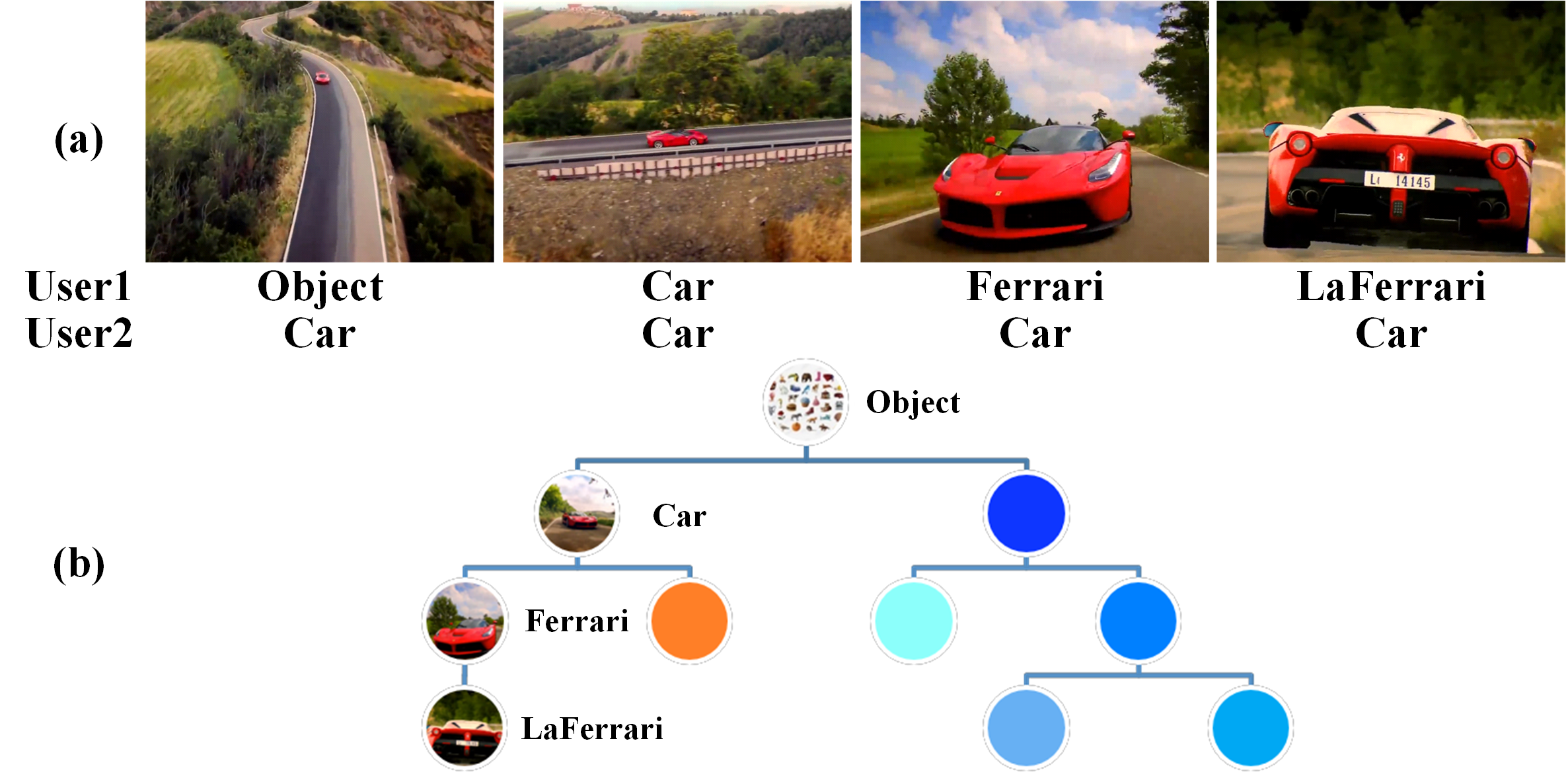}
	\caption{(a) Examples of annotations by different users on a set of images with the same object in different scenarios. (b) Visual data can be organised as a visual hierarchy,  like the lexical hierarchy that is organised based on linguistic descriptions. \label{fig:1}}
\end{figure}

However, visual data in many datasets usually get subjective labels, as 1) the same object in the different scenarios may be given different representations, 2) annotators may have different understandings of the same visual data. An observation is that people usually have different linguistic descriptions of an object. As shown in figure \ref{fig:1} (a), suppose we see something moving towards us at a certain distance. At first, we do not know what it is. As it gets closer, we realise that it is a car. As it gets closer to us, we can see signs on the car telling us that it is a Ferrari. When it gets close enough, someone will recognise it as a LaFerrari on the basis of its appearance. Thus, for one object in different scenarios, it may get four labels. Another case is that someone may call all four images as "Car", since they have no idea about "Ferrari" and "LaFerrari". How the images are annotated depends on the understanding and knowledge of the annotated. This is what we call the semantic gap problem (SGP)\cite{23}. The semantic gap is defined as the difference between the linguistic representations of the same concept, which usually leads to the misunderstanding between two individuals with different knowledge backgrounds. Since images annotated by natural language are usually used as training and test set in current computer vision tasks, there is an inevitable bias between image descriptions caused by SGP, resulting in poor performance of trained models.

To address the above issues, we propose a novel unsupervised visual hierarchy construction method to build visual semantics, aiming to build a classification model by learning from pure-visual information and to dissipate the bias of linguistic representations caused by SGP. 
Our intuition comes from real-world knowledge representation, the visual information of concepts can be hierarchically organised, where each concept can be described by a set of features rather than a linguistic annotation, as shown in Fig \ref{fig:1} (b).  The visual hierarchy conveys the uniform visual expressing for the substance concepts, which we call visual semantics.
The visual hierarchy expressing visual semantics is learned autonomously by our proposed algorithm. In this process, we model object categories as substance concepts, which are organised into a visual hierarchy; objects in the visual data are modelled as visual objects that are stored in each node of the visual hierarchy as a set of visual features like properties.
The evaluation consists of two parts, firstly we apply the constructed hierarchy on the object recognition task and then we compare our visual hierarchy and existing lexical hierarchies to show the validity of our method. It also provides the possibility to transform the visual hierarchy in substance concepts into a lexical hierarchy that humans can understand. 

\section{Method}
\label{sec:Method}
In this section, we describe how to organize visual data to represent substance concepts. As demonstrated in real-world knowledge representations, taxonomies usually organise objects in a hierarchy, which provide references for visual semantics to represent substance concepts. To build visual semantics, we propose to organize visual data as a visual hierarchy, where each node as a concept can be described by a set of features. To this end, we design a novel unsupervised visual hierarchy construction method, aiming to build a classification model by learning from pure-visual information and to dissipate the bias of linguistic representations caused by SGP. 

The visual hierarchy construction algorithm is inspired by \cite{67}. The algorithm for building the visual hierarchy that introduces two pre-defined properties that represent object features, called \textit{Genus} and \textit{Differentia}, instead of the original labels of visual data, for training. \textit{Genus} represents a set of previously defined properties that an object satisfies and is used as part of the definition of a new object; \textit{Differentia} is a new set of properties that are used to differentiate among objects with the same genus. \textit{Genus} and \textit{Differentia} are applied to determine the specific location of a visual object in the hierarchy, including whether it is an existing concept, a new concept, or needs to be iteratively determined.

Note that memory in the algorithm is a collection of concepts, which in turn are collections of visual objects. The collection of concepts in the memory is organized in a hierarchy of nodes, where each child node presents all the features of the father node plus some more details that distinguish the node itself from the father and siblings. All concepts in nodes are defined by appropriate visual objects or genus objects, where the visual data is stored as feature maps.

To evaluate the constructed visual hierarchy, we apply it to an object recognition task. Furthermore, we construct a lexical hierarchy based on the super-subordination relation among synsets in the large lexical database WordNet \cite{75}, which is compared with our visual hierarchies to demonstrate the validity of our method.

\section{Results and Future Work}
\label{sec:Conclusion}
\subsection{Preliminary results}

To learn features of the visual data, we apply VGG-16 \cite{85} 
for pre-training on ImageNet \cite{26} 
and grouped the videos into similar frames with DeeperCluster \cite{84}. To complete the experiments in building the hierarchy, we explored the advantages of ground truth based on visual Genus and Differentia by reorganising and annotating the visual data. When trained with the same depth models and several datasets with different ground truths, our dataset achieves a relative performance improvement of around 10\% on all depth models. \cite{1} reports the details and results of our experiments.

\subsection{Future work}

Our project is well underway now. In this project, we make efforts to build a classification model by learning from pure-visual information and to dissipate the bias of linguistic representations caused by SGP. Preliminary experimental results have demonstrated its superiority, while the success of the bag-of-words model \cite{zhang2010understanding} and the data distillation model \cite{wang2018dataset} has provided us with viable solutions for feature analysis on visual object hierarchy.

The visual object hierarchy we obtained can also be applied to several tasks, including object recognition, content-based image retrieval, scene understanding, etc. Besides, the visual object hierarchy will also play an important role in machine learning interpretability by providing a macroscopic view of how a machine learning model accomplishes a task. 
It will provide a solution for explaining how the machine works and it also offers the possibility for humans to trust machine decisions in their future work, which proves to be a promising topic.

\bibliographystyle{named}
\bibliography{ijcai22}

\end{document}